\documentclass[10pt,twocolumn,letterpaper]{article}

\usepackage{times}
\usepackage{epsfig}
\usepackage{graphicx}
\usepackage{amsmath}
\usepackage{multirow}
\usepackage{amssymb}
\usepackage[numbers,sort]{natbib}
\usepackage[breaklinks=true,bookmarks=false]{hyperref}

\date{\vspace{-5ex}}

\begin{document}


\title{Signs in time: Encoding human motion as a temporal image}

\author{Joon Son Chung, Andrew Zisserman\\
Visual Geometry Group, Department of Engineering Science\\
University of Oxford\\
{\tt\small (joon,az)@robots.ox.ac.uk}
}

\maketitle


\begin{abstract}
The goal of this work is to \textit{recognise} and \textit{localise}
short temporal signals in image time series, where strong supervision is
not available for training.

To this end we propose an image encoding that concisely represents
human motion in a video sequence in a form that is suitable for
learning with a ConvNet. The encoding reduces the pose information from
an image to a single column, dramatically diminishing the input requirements
for the network, but retaining the essential information for recognition.

The encoding is applied to the task of recognizing and localising
signed gestures in British Sign Language (BSL) videos.  We demonstrate
that using the proposed encoding, signs as short as 10 frames duration
can be learnt from clips lasting hundreds of frames using only weak
(clip level) supervision and with considerable label noise.
\end{abstract}


\vspace*{-2mm}
\section{Introduction}
\vspace*{-1mm}

One of the remarkable properties of deep learning with ConvNets is
their ability to learn to classify images on their content given only
image supervision at the class level, \textit{i.e.}\ without having to provide
stronger supervisory information such as bounding boxes or pixel-wise
segmentation. In particular the position and size of objects is
unknown in the training images. This ability is evident from the
results of the ImageNet and PASCAL VOC classification 
challenges. Furthermore,
several recent works have also shown that given only this class-level 
image weak-supervision, the trained networks can to some
extent infer the localization of the objects that the image 
contains~\cite{Oquab15,Papandreou15,Simonyan14a}.

In this paper we take advantage of this ability to recognize temporal
signals in an image time series. Our
aim is to obtain ConvNets that can both {\em classify} a video clip
as to whether it contains a target sequence or not, and {\em
localize} the target sequence in the clip, using only class level
supervision of the clip. 
Why is this challenging? There are two reasons, first we
consider target sequences that are very short within clips that are
long -- for example a target lasting less than 10 frames in a clip of
hundreds of frames (a target less than 0.5s in a 12s clip); second, the
supervision can be not only weak, but also noisy. 

To achieve this we propose a novel encoding of human motion in a video
sequence that concisely represents the framewise human pose information in a 
manner that can be utitlized by a ConvNet. For example, 10 seconds of video
is condensed to a 250 (i.e.\ $25 \times 10$) pixel width image, with 
a height of only 10 pixels.

We apply this representation to the task of recognizing gestures
(signs) in British Sign Language, where the provided supervision is
both weak and noisy.  The outcome of using this encoding is that it is
possible to learn and localize short temporal hand gestures in long
temporal clips, that are virtually invisible to a non-expert. This is
a `needle in a haystack' problem, where the needle is unknown.  Note,
it is a {\em sequence} that must be recognized -- the target cannot be
spotted in a single frame or time instance (as is the case for some
human actions, \textit{e.g.}\ playing an instrument).

To the best of our knowledge, this is the first time
ConvNets have been used to recognise and localise complex temporal
sequences, such as the gestures in sign language,  in a video sequence
using such weak and noisy annotation in training.
The performance far exceeds previous work in this area in terms
of supervisory requirements and generalization across signers.


\vspace{-10pt}\section{Encoding motion}\vspace{-6pt}
\label{sec:findtime}

Given a video clip of sign language
gestures, the objectives are to determine if a target sequence is
present in the clip, and, if so, where it is. 
There are two key research questions: 
(i) how to encode the image time series, and (ii) the 
design of the ConvNet architecture to recognize the target sequence. Of course,
these two issues are coupled. 

The data to be represented consists of the pixel coordinates of the
two hands and head in each frame, hereafter referred to as keypoints,
i.e.\ six values. Details of how these points are obtained are given
in Section~\ref{sec:details}.

\begin{figure}[ht]
	\centering 
	\vspace{-8pt}
	\includegraphics[width=0.45\textwidth]{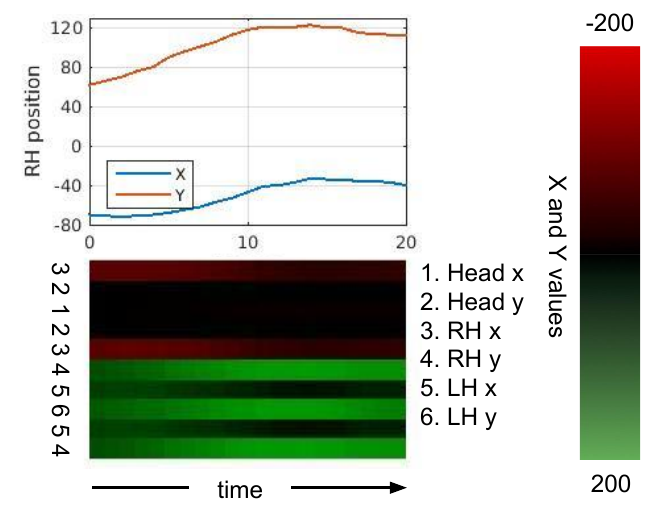}
	\vspace{-8pt}
	\caption{\small{{\bf The {\em kinetogram} temporal encoding of human keypoint motion}.
The two
channels representing position are shown, with red representing
negative values and green positive.  The represented motion is for the
BSL sign for \textit{`valley'} (a V-shape drawn with both hands moving downwards). Note
that row 3 representing the right-hand (left in the image) \textit{x}-values get dimmer with
time as $x$ increases and the values become less negative, whereas row 4 representing
\textit{y}-values get brighter with time as $y$ increases and the values become more positive.
The representation is reflected vertically (at 1 and 6) to mitigate boundary effects for the filters.
}}
\label{fig:input_heatmap} 
\vspace{-5pt} \end{figure}

The key idea of the encoding is to represent the six values for each frame
as intensity values in a column of a heatmap-like 
image, using two bytes per value.
This is simply equivalent to treating
the matrix of vectors of keypoint positions against time as an image. 
The velocity (frame difference in position) of the
keypoints is also encoded in a further channel, storing the values in a 
heatmap, as for the position. 
In summary, two bytes (the first two channels) are used to store the position,
which requires more precision, and one byte (the third channel) is
used for the velocity (the frame difference in position).  
Figure~\ref{fig:input_heatmap} shows an example of the encoding, which we term a
{\em kinetogram}.
In this
representation, an upward motion gives a decrease in the brightness in
the row relating to the body joint (as the $y$ value reduces) .

\noindent\textbf{Discussion.} 
This motion encoding was chosen as one that should be suited to
convolutional filter learning. For example, horizontal temporal
derivative filters on the brightness values can measure if the hand is
moving upwards (negative output) or is stationary (zero
output). Filters covering several rows can detect if the hands are
moving together or not, etc.  This encoding has the properties of being
compact and minimal. We did consider several other
representations, but rejected these as they resulted in much larger
input images. For example, aside from simply using all the frames of the clip
(e.g.\ 300) as input channels, the motion could be encoded as
 optical flow
in the manner of~\cite{Simonyan14b}, but that would require two images
per frame (one image for each of the horizontal and vertical components), 
even if only the motion of the keypoints was recorded in each.
A second possibility is to build Motion History
Images~\cite{ahad2012motion} or its more modern incantation~\cite{Bilen16a};
but in this case the background
motion would be extremely distracting and challenging 
(see Figure~\ref{fig:trackloc}d, the signer is overlaid on the original
broadcast video).

\subsection{ConvNet architecture} 
\begin{figure}[ht]
\centering 
\vspace{-10pt}
\includegraphics[width=0.35\textwidth]{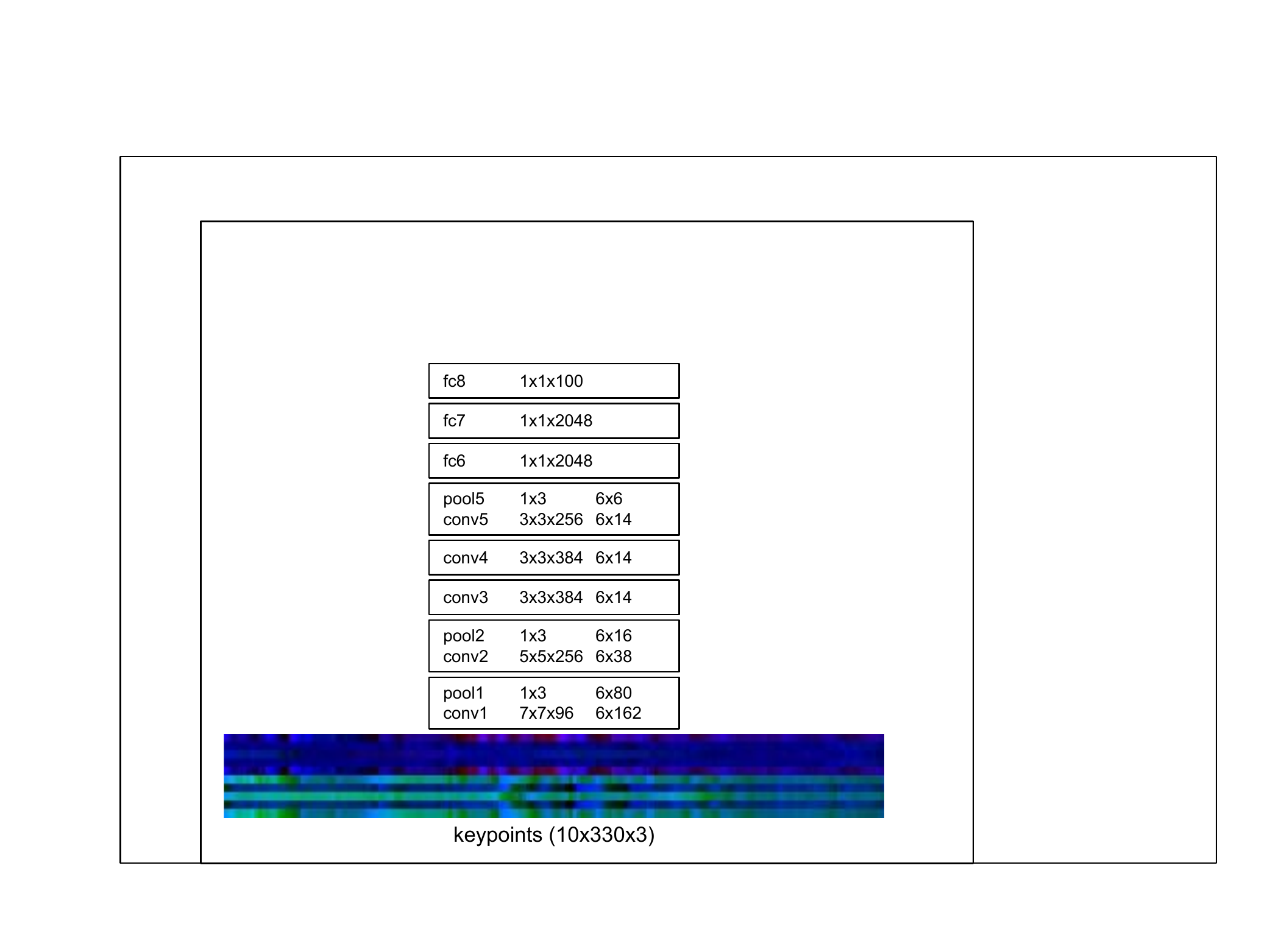}
\vspace{-10pt}
\caption{\small{{\bf ConvNet architecture.} Max-pooling is used in all pooling layers.
The five numbers following each convolutional layer specify: the size of the filters, the number of channels, and the resolution of the layer.
}}
\label{fig:arc} 
\vspace{-5pt} 
\end{figure}

We use a convolutional neural network inspired by those designed for
image recognition. Our layer architecture (Figure~\ref{fig:arc})
is based on AlexNet~\cite{Krizhevsky12}, but with
modifications. AlexNet takes a square image of size 224$\times$224 pixels,
whereas our input size is at least 330 pixels (the number of time steps) in
the time-direction, and only 10 pixels in the other direction (so the input
image is $10 \times 330$ pixels). 

\vspace{-10pt}\subsection{Localisation via backprop}\vspace{-5pt}

\begin{figure}[ht]
\vspace{-10pt}
\centering 
\includegraphics[width=0.48\textwidth]{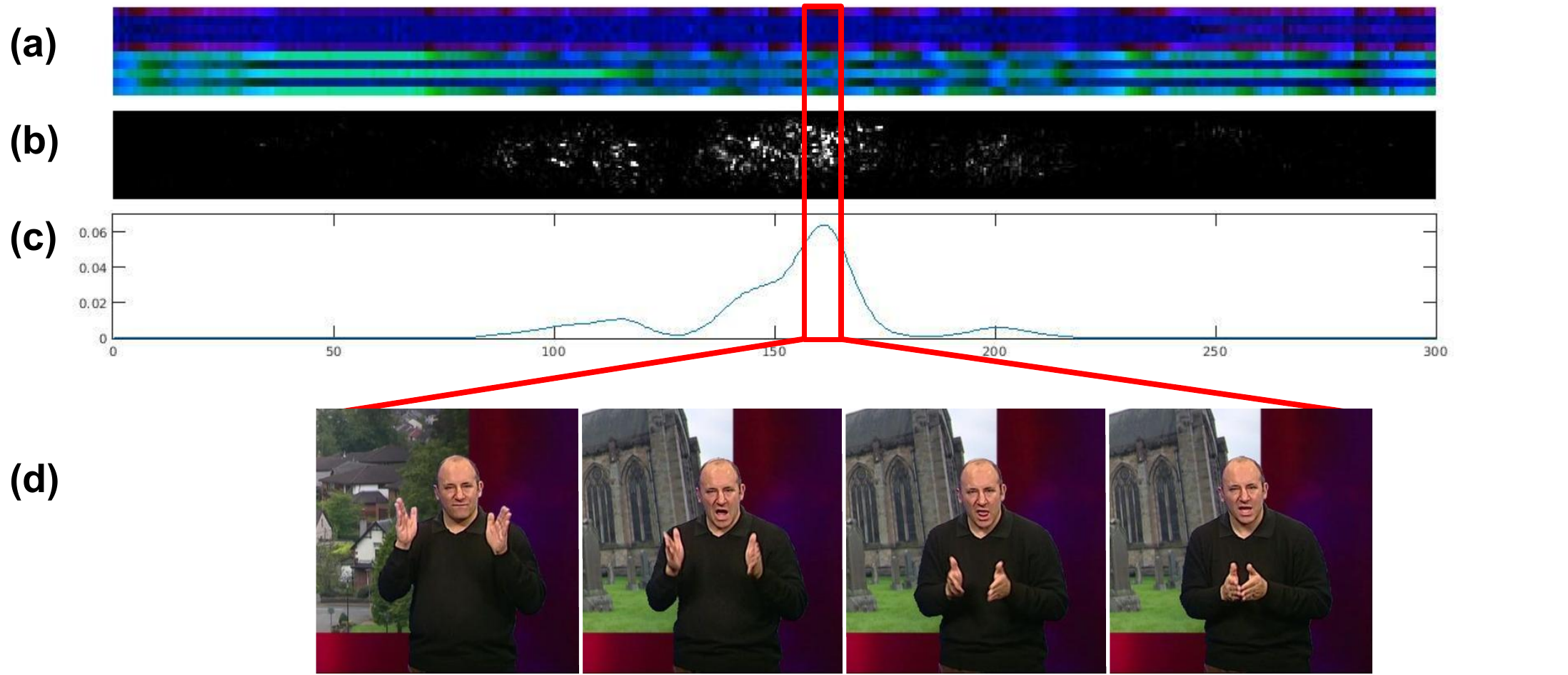}
\vspace{-15pt}
\caption{\small{Salient region for `\textit{valley}', localised only using motion information. The sign for `\textit{valley}' is a large V drawn with both hands. (a) Kinetogram; (b) Saliency map; (c) Saliency over time, after Gaussian filtering; (d) The corresponding target sequence.} }
\vspace{-5pt}
\label{fig:trackloc}
 \end{figure}

The objective here is, once the networks have been trained to classify
the clip, localize the target sequence within (positive) clips. 
Simonyan~\textit{et al.}~\cite{Simonyan14a} have shown that it is
possible to infer the localization of visual objects in an image
 as a saliency map for a
network trained to classify images. We adapt this method
to time series to find the salient temporal intervals in the input signal that
have high influence on the class score. 

The method proposes that the partial derivative 
$\left. w = \frac{\partial S_c}{\partial I}\right\rvert_{I_0}$ 
approximates the contribution that individual
pixels make to the class score.
This derivative is obtained by 
back-propagating from the
class score $S_0(I_0)$ to the image.

In our case, the derivative $w$ shares the dimensions of the input
time series.  We compute saliency $M_{i,j}$ at position $i,j$ as
$M_{i,j} = \max_{c}
\rvert w_{i,j,c} \rvert$ (\textit{i.e.} the channel-wise maximum over every pixel, the result is shown in 
Figure~\ref{fig:trackloc}b). A 2-dimensional Gaussian is used to smooth the signal, and then the column-sum
is taken to obtain a score function against time (Figure~\ref{fig:trackloc}c).

	
	\vspace{-10pt}\section{Dataset}\vspace{-6pt}
\label{sec:datasets}

\begin{table}[ht]
\vspace{-8pt}
\centering
\scriptsize \begin{tabular}{| l | r | r |}
  \hline
  \textbf{Label} & \textbf{Single} & \textbf{Multiple} \\ \hline 
  Total \# of programmes & 890 & 890 \\ \hline 
  Total video length (hours) & 678 & 678 \\ \hline 
  Vocabulary size & 100  & 100 \\ \hline 
  Total \# of subtitles & 662,165 & 662,165 \\ \hline 
  Useful \# of subtitles & 50,000 & 104,247 \\ \hline 
  Min./max. instances per class & 500/500 & 500/2000 \\ \hline 
  \# of words per subtitle & 9.96 & 10.11 \\ \hline 
  \textit{In-vocab} words per subtitle & 1.00 & 1.22 \\ \hline  
\end{tabular}
\vspace{-8pt}
\caption{\small{Dataset statistics -- `Multiple' has multiple words from the vocabulary in a 
clip, and `Single'
(a subset of `Multiple') only has a single word from the vocabulary in each clip}}
\label{table:dataset}
\vspace{-8pt}
\end{table}

We collect a new dataset for this task that is used
for recognition and localisation of
signed gestures. The dataset consists of 890
high-definition `\textit{sign-interpreted}' TV broadcast videos 
aired between 2010 and 2016. We use the video and the
corresponding subtitles as the weakly labelled training data for the
tasks. 
The format of the dataset is
similar to that of~\cite{Pfister13}, but orders of magnitude larger in
scale.
Table~\ref{table:dataset} shows the key statistics of the
dataset.
A vocabulary of 100 target words are selected
primarily based on their frequency of appearance in the programmes. Stop
words and words with more than
one meaning such as `\textit{match}' and `\textit{bank}' are excluded
from selection. We also selected programmes primarily on the
genres of `wildlife' and `cooking',
in order to reduce this polysemy problem.

We select two datasets: one has multiple words from the vocabulary in a 
clip, and the other
(a subset of the first) only has a single word from the vocabulary in each clip.
The first results in multiple labels per clip for training. This
is beneficial for 
two reasons: (i) we eliminate the need to
discard training sequences that belong to multiple classes, hence
increasing the amount of training data available. (ii) it
improves the ratio of supervision per subtitle in the training data, 
in our case, by a factor of 1.22.

A sequence is extracted for each occurrence of the target word in the subtitles. The alignment between the subtitle and the signs is imprecise, therefore the temporal window is padded by an additional 8 seconds. The total length of each training sequence is over 300 frames (12 seconds), whereas most subtitles are shorter than 4 seconds.

The dataset is divided into training, validation and test subsets (80:10:10) in chronological order, the test set being the oldest.

\noindent\textbf{Discussion.} 
This dataset is particularly challenging for a number of reasons: (i)
the word order in the subtitle is not the same as the order in which
they are signed, and furthermore the alignment between the sign and the
subtitle is unknown and the offset can be more than 5 seconds. Hence
we cannot estimate when the word might be signed; (ii) a word that
appears in the subtitle may not be signed (the proportion of signed
video which actually contains the target word is only 20-60\%,
depending on the word); (iii) the contents are signed by 50 different
signers; and finally, (iv) there is a large variation in content, from
`cooking' to `wildlife', broadcast over a period of 6 years.


\vspace{-10pt}\section{Implementation details} 
\label{sec:details}

\vspace{-2.5pt}\subsection{Data preparation}\vspace{-2.5pt}

\noindent\textbf{Text extraction and processing.}
British TV transmits subtitles as bitmaps rather than as text, therefore subtitle text is extracted from the broadcast video using standard OCR methods~\cite{Buehler09}. Subtitles are stemmed (\textit{e.g.} `\textit{played}', `\textit{played}', `\textit{playing}' all become `\textit{play}') and stop words (\textit{e.g.} `\textit{a}', `\textit{the}') are removed.

\noindent\textbf{Upper-body tracking.}
 We use the ConvNet-based upper body pose estimator of~\cite{Pfister15a} to track the head, elbows and hands of the signer. The input to the tracker is a crop of the signer around 900 $\times$ 900 pixels, from a Full HD (1920 $\times$ 1080) frames. The pose estimator generates a confidence score for each keypoint, and one usually takes the maximum to estimate the location of the keypoint. However, the returned confidence heatmaps for some keypoints often have a multi-modal distribution (\textit{e.g.} the left-hand detector gives high confidence for both hands), which can give incorrect estimates. Dynamic programming in time corrects many of these errors by optimising between the framewise confidence and the distance of the keypoints between neighbouring frames. This improves the tracking performance from 95.7\% to 97.6\% (PCKh-0.5) on the wrist.

\vspace{-2.5pt}\subsection{Training}\vspace{-2.5pt}
\label{sec:training}

\noindent\textbf{Loss functions.} 
We use the {\em weighted binary logistic loss}, for a binary
classification (present/ not present) for each class. The loss is
weighted to deal with imbalance in the training data:
 $L(S,l) = w_{l} \log{ (1 + \exp(- l S)) }$, where $S$ is the class score (\textit{fc8} output), $l$ is the binary class label (present/ not present) and $w_{l}$ is the ratio $n_{neg} / n_{pos}$ for each class when $l=1$, and $1$ when $l=-1$.
	 
\noindent\textbf{Data augmentation.} 
There are three augmentation steps:
The video is played back at three
different speeds, for which the velocities must be recomputed; the
coordinates of the keypoints (the tracker output) are randomly
shifted; and the brightness of the kinetogram image is also varied, which is
equivalent to spatially scaling the input video.

	\noindent\textbf{Details.}
	Our implementation is based on the MATLAB toolbox MatConvNet. The network is trained with batch normalisation. Despite this, a slow learning rate of $10^{-3}$ to $10^{-4}$ was used to get a stable learning, due to the label noise.


\vspace{-10pt}\section{Experiments}\vspace{-6pt}
\label{sec:experiments}

In the following experiments the network is trained on the dataset of
Section~\ref{sec:datasets}, and the results are excellent -- as can be
seen qualitatively in the accompanying video ({\tt\small https://youtu.be/ujQaRPIlexQ}).

However, the noise in the supervision
(that only 20--60\% of the words in the subtitle are actually signed)
presents a problem for quantitative 
evaluation as even a perfect classifier would
not score well under such circumstances. We deal with this problem by
giving results on an external test set~\cite{Pfister14a} for which the
labelling is not noisy.

\noindent\textbf{External test dataset.} The test dataset is based on the BBC sign language videos of \cite{Pfister14a}. This dataset is independent from our main dataset, and the format is the same as the data used by~\cite{Pfister13,Pfister14}, which makes it useful for comparisons. A number of words that appear frequently both in our training dataset and in the external test set (see Table~\ref{table:lossresults}) have been manually annotated at frame-level, which is used to evaluate both the classification and localisation tasks.

\begin{table}[h] 
\vspace{-5pt}
\setlength{\tabcolsep}{1.5pt}
\centering 
\small
\begin{tabular}{ l l l l l l l l l l l l}

\textbf{\textit{Label}} & \rotatebox{90}{beef} & \rotatebox{90}{chocolate} & \rotatebox{90}{milk}  & \rotatebox{90}{pig} & \rotatebox{90}{rain} & \rotatebox{90}{school} & \rotatebox{90}{soup}  & \rotatebox{90}{valley} & \rotatebox{90}{war} & \rotatebox{90}{winter} & \rotatebox{90}{\textit{\textbf{mAP}}} \\  \hline

  Single~ 		& 43.7 & 45.4 & 44.7 & \textbf{51.4} & 76.6 & 27.8 & 18.5 & 18.9 & 49.7 & 61.7 & 43.9 \\ 
   Multi~ 		& \textbf{59.0} & \textbf{70.4} & \textbf{58.8} & 49.5 & \textbf{80.6} & \textbf{48.9} & \textbf{62.3} & \textbf{49.1} & \textbf{67.9} & \textbf{80.8} & \textbf{62.7} \\ \hline \\

\end{tabular}
\vspace{-20pt}
\caption{\small{Average Precision for \textbf{gesture localisation} on the external BBC test set.}}
\label{table:lossresults}
\end{table}

\noindent\textbf{Evaluation protocol.} 
The task is to localise the temporal interval of the half-second target gesture within the 12-second window and provide a ranked list of temporal windows in the order of confidence. If the gesture overlaps at 50\% with the ground truth, the localisation is deemed successful.

\noindent\textbf{Localisation results.}
The words that appear frequently in our dataset are different from those of~\cite{Pfister13} and~\cite{Pfister14}; therefore we must compare the performance figures with caution. Our test set annotation methods and the evaluation protocol closely follow that of~\cite{Pfister13} and~\cite{Pfister14}. Comparing our performance figures to Figure 7 of~\cite{Pfister14}, it is clear that our localisation performance is competitive with the strongly supervised method of~\cite{Pfister14} (which uses a dictionary) and far exceeds the previous best weakly supervised method of~\cite{Pfister13}. For example, our average precision on \textit{`winter'} (which appears in both our work and theirs) is 81\%,~\cite{Pfister14} is 50\% and~\cite{Pfister13} is 18\% 
(note, the performance figures of~\cite{Pfister13,Pfister14} are not available, so values are
estimated from the graphs). 
There are 6 words (\small{beef, chocolate, jelly, milk, war, winter}\normalsize) that appear in common between our dataset and~\cite{Pfister13}. Our mAP over these words are 62\%. This compares to~\cite{Pfister13}'s mAP of 17.8\% when the motion input (same modality as ours) is used, 
and 57.1\% for the multi-modal case where the hand shape and the mouthing is used as well. 
The other words in our evaluation do not
appear in~\cite{Pfister13,Pfister14}, but the performance figures are
competitive with those that do. It is notable that the performance of strongly supervised
methods can be matched, particularly when the network has never been
explicitly trained to localise these signals.


\vspace*{-2mm}
\section{Conclusions}
\vspace*{-1mm}

We have demonstrated that a {\em kinetogram} encoding of human motion in
combination with a standard ConvNet is a very powerful representation
and learning machine. And we have shown its use in recognizing gestures
in sign language using training with weak and noisy supervision.

More
generally, the encoding is applicable to other situations that involve
plucking sequences out of long clips. 
For example, keyword spotting in always on speech recognition, identifying pathologies in medical time series data, or localizing 
human actions in videos.

{\footnotesize
\bibliographystyle{splncs03}
\bibliography{signsintime}
}

\end{document}